\documentclass[11pt,letterpaper]{article}
\usepackage{emnlp2016}
\usepackage{times}
\usepackage{latexsym}
\usepackage{graphicx}
\usepackage{color}
\usepackage{tabularx}

\emnlpfinalcopy
\def\emnlppaperid{XXXX}

\newcolumntype{Y}{>{\centering\arraybackslash}X}

\newcommand{\refexp}[1]{\emph{``#1''}}

\title{Utilizing Large Scale Vision and Text Datasets for Image Segmentation from Referring Expressions}

\author{Ronghang Hu$^1$ \and Marcus Rohrbach$^{1,2}$ \and Subhashini Venugopalan$^3$ \and Trevor Darrell$^1$ \\
$^1$UC Berkeley \qquad
$^2$ICSI, Berkeley \qquad
$^3$UT Austin \qquad
}

\date{}

\begin{document}

\maketitle

\begin{abstract}
Image segmentation from referring expressions is a joint vision and language modeling task, where the input is an image and a textual expression describing a particular region in the image; and the goal is to localize and segment the specific image region based on the given expression. One major difficulty to train such language-based image segmentation systems is the lack of datasets with joint vision and text annotations. Although existing vision datasets such as MS COCO provide image captions, there are few datasets with region-level textual annotations for images, and these are often smaller in scale. In this paper, we explore how existing large scale vision-only and text-only datasets can be utilized to train models for image segmentation from referring expressions. We propose a method to address this problem, and show in experiments that our method can help this joint vision and language modeling task with vision-only and text-only data and outperforms previous results.
\end{abstract}

\section{Introduction}

\begin{figure}[t]
\centering
\includegraphics[width=\linewidth]{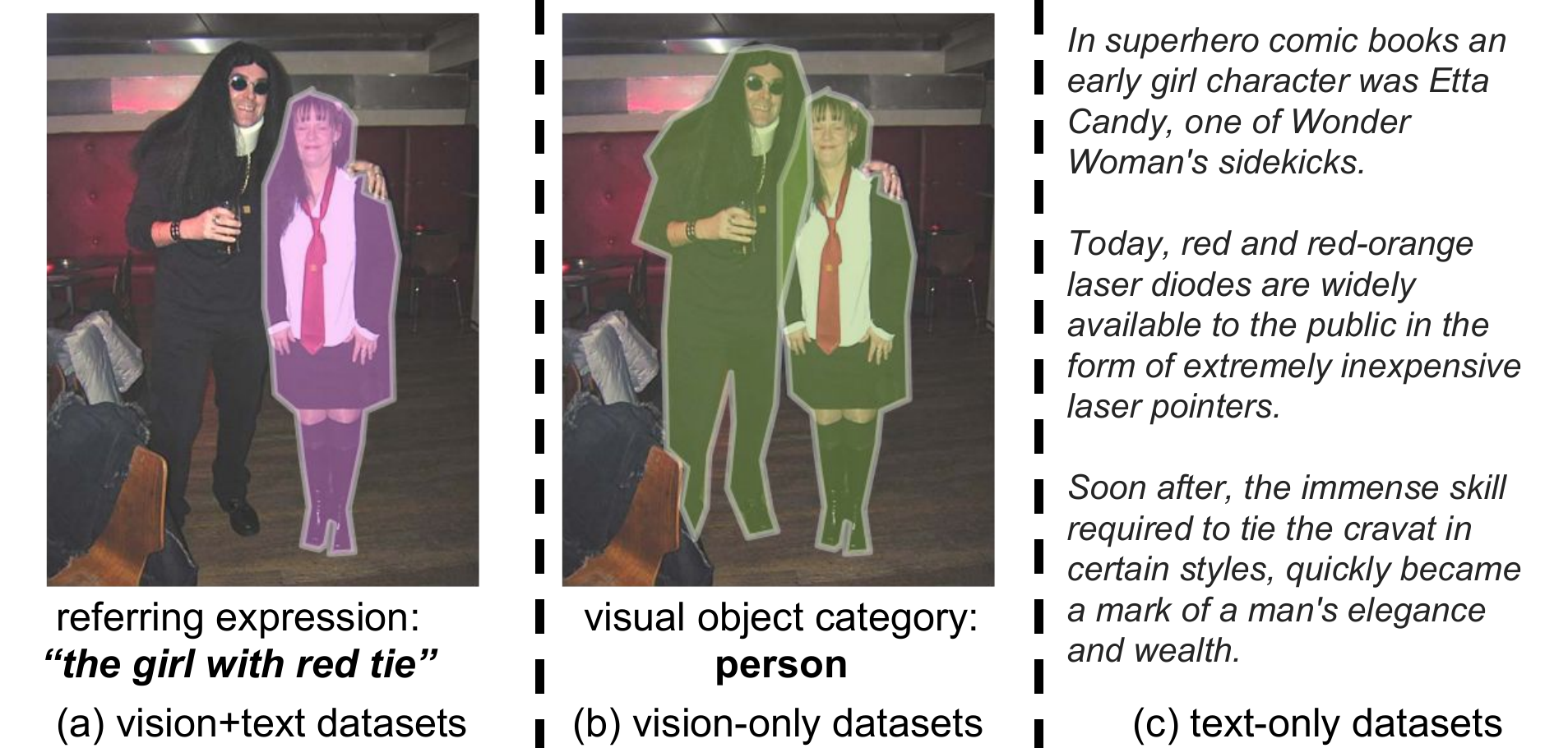}
\caption{Previous methods for image segmentation from referring expressions require joint vision and text annotations (shown in a), but such datasets are expensive to collect. We explore how datasets with vision-only (shown in b) and text-only (shown in c) annotations can be utilized in this task. }
\label{fig:teaser}
\end{figure}

Semantic image segmentation \cite{shotton2009textonboost,long2015fully,chen2015semantic,zheng2015conditional} is an important problem in computer vision. Given an input image and a pre-defined set of visual semantic categories, such as ``sky'', ``dog'', ``bus'', the task of semantic image segmentation is to localize all image pixels that belong to a particular category.

Instead of operating over a fixed set of visual categories, recent works such as image captioning \cite{mao2015deep,donahue2015long,devlin2015language} and visual question answering \cite{xu2015ask,yang2016stacked,andreas2016learning} have extended visual comprehension from a set of classes to broader semantic labels represented by natural language expressions. \newcite{hu2016segmentation} approaches the task of \textit{image segmentation from referring expressions}, where the goal is to ground a given query expression in an input image, and output a pixelwise segmentation for the corresponding visual entity described by the expression. For example, given an image and an expression ``the girl with red tie'' (Fig. \ref{fig:teaser}a), the model is asked to output pixelwise segmentation mask for the girl on the right.

\newcite{hu2016segmentation} proposes a model that encodes the given expression into a real-valued vector using Long Short Term Memory (LSTM) networks \cite{hochreiter1997long} and extracts a spatial feature map from the image using a Convolutional Neural Network (CNN) \cite{krizhevsky2012imagenet}. Then it performs pixelwise classification based on the encoded expression and feature map to output an image mask covering the visual entity described by the expression.
However, the LSTM-CNN model proposed in \newcite{hu2016segmentation} requires referring expression annotations at image region level as training data (Fig. \ref{fig:teaser}a). Such annotations are expensive compared with visual class labels, and existing image segmentation datasets with referring expression annotations \cite{kazemzadeh2014referitgame,mao2016generation} are an order of magnitude smaller than those with only visual category annotations such as MS COCO \cite{lin2014microsoft}.

Therefore, in this paper, we propose a method to utilize existing large scale vision-only datasets containing image regions annotated with visual class labels but no text (Fig. \ref{fig:teaser}b), and text-only corpus datasets (Fig. \ref{fig:teaser}c) to help image segmentation from referring expressions. We show that the performance of this task can be improved by pretraining word embeddings on text corpus and synthesizing textual phrases from the class names of visual classes as additional training data. We also incorporate traditional category-based semantic image segmentation datasets and models by mapping the textual expression into visual categories and matching it with category-based image segmentation results.

Our work is related to bounding box or pixelwise image region localization from query expressions \cite{hu2016natural,mao2016generation,rohrbach2015grounding} and image captioning with vision-only and text-only data \cite{hendricks2016deep}.

\section{Our Method}
\label{sec:method}

Our method extends the LSTM-CNN model \cite{hu2016segmentation} with word embeddings (Sec. \ref{sec:method_1}) and synthesized expressions (Sec. \ref{sec:syn_expr}), and exploits category-based image segmentation datasets and models (Sec. \ref{sec:method_2}). In Sec. \ref{sec:all-in-1} we describe how we jointly train the full model.

\subsection{Word Embeddings}
\label{sec:method_1}

\begin{figure}[t]
\centering
\includegraphics[width=0.9\linewidth]{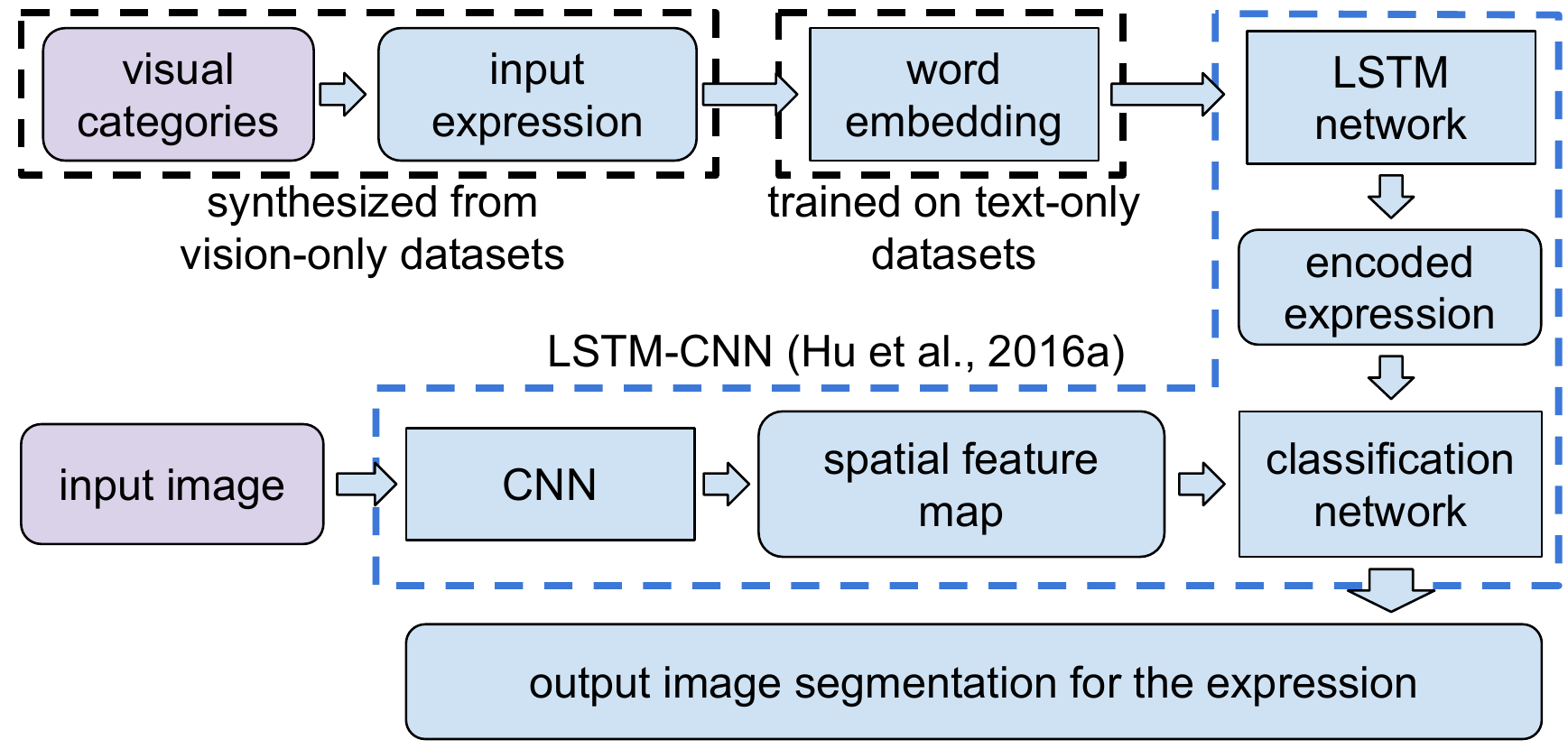}
\caption{We augment the previous LSTM-CNN model with a word embedding pretrained on text-only corpora datasets, and synthesized textual expressions as additional training data from vision-only semantic image segmentation datasets. }
\label{fig:method_1}
\end{figure}

To utilize existing text-only datasets, we train a word embedding matrix on a large text corpus using GloVe \cite{pennington2014glove}, which is effective in embedding novel and rare words. The trained GloVe vectors are used as the word embedding matrix transforming the inputs to the LSTM network of \cite{hu2016segmentation} as shown in Fig. \ref{fig:method_1}. This matrix is kept fixed during image segmentation training.

Compared with learning a word-to-vector mapping from scratch on the limited joint vision and language data as in \newcite{hu2016segmentation}, the GloVe vectors trained on large corpora are more effective in projecting the words to a semantic space, and handling rare and novel words not seen in the training set.

\subsection{Synthesized Expressions}
\label{sec:syn_expr}
The large scale vision-only datasets for semantic image segmentation, such as MS COCO \cite{lin2014microsoft}, only have image regions annotated over a pre-defined set of visual categories. To utilize such vision-only datasets, we synthesize textual expressions from the category annotations. In our implementation, we take a simple approach and directly use the visual class name as the textual expression for a image region, as shown in Fig. \ref{fig:method_1}. For example, an image region of the visual class \textit{person} is labeled with expression ``person''. These synthesized expressions and the image regions are used as additional training data for image segmentation from referring expressions. This approach, while straightforward, has the potential to also benefit semantically related words. Since the GloVe vectors map semantically related words to close points in the projection space, an image region of \textit{person} class can also benefit expressions with semantically related words such as ``man'', ``girl'' and ``child''.

\subsection{Category-based Image Segmentation}
\label{sec:method_2}

\begin{figure}[t]
\centering
\includegraphics[width=0.9\linewidth]{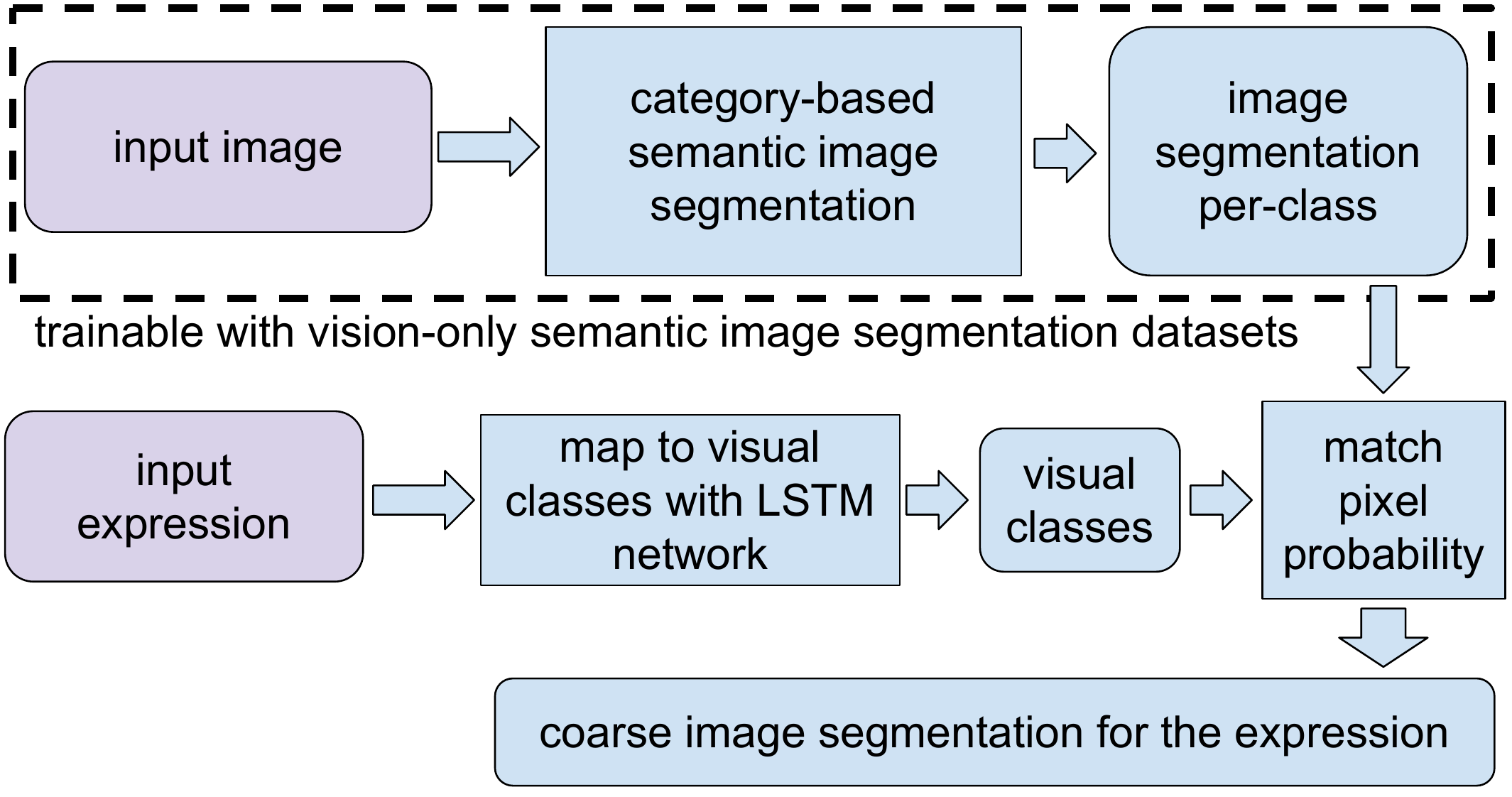}
\caption{To utilize existing category-based semantic image segmentation models, we map the input expression into visual classes with a LSTM classifier, and match the mapped classes to category-based image segmentation results to obtain a coarse image segmentation output. }
\label{fig:method_2}
\end{figure}

In the computer vision community, the traditional category-based semantic image segmentation problem has been extensively studied, and a number of state-of-the-art models for this task have been proposed \cite{long2015fully,chen2015semantic,zheng2015conditional}. In category-based image segmentation, a classification model is trained on vision-only datasets (Fig. \ref{fig:teaser}, b) to label every pixel with a visual class from a pre-defined set of $M$ visual classes. For each pixel $(i,j)$ in the image, a probability vector $\mathbf{p}^{(i,j)}_{image}$ is outputted over the $M$ classes.

To take the advantage of the well-studied category-based image segmentation models, we also associate the input textual expression with the visual categories by classifying the expression into the $M$ classes as shown in Fig. \ref{fig:method_2}. For example, the expression ``a kitty sitting at the table'' should be associated with \textit{cat} class. We treat the expression as a word sequence and feed it into a LSTM network. After scanning through the sequence, a two-layer neural network takes the LSTM final hidden state and outputs a $M$-class probability vector $\mathbf{p}_{text}$.

Then, the category probability distribution at each image pixel is matched with the probability distribution of the textual expression. If both an image pixel and the textual expression have $0.99$ probability to be class \textit{cat}, then it is likely that this pixel belongs to the image region described by the expression (assuming only one cat present in the image). We define the foreground probability $p_{fg}(i,j)$ of pixel $(i,j)$ belonging to the input expression as the dot product between the two probability vectors. 
\begin{equation}
p_{fg}(i,j) = <\mathbf{p}^{(i,j)}_{image}, \mathbf{p}_{text}>
\end{equation}

If both $\mathbf{p}^{(i,j)}_{image}$ and $\mathbf{p}_{text}$ are accurate, then this approach should be able to get the visual category correct in the output segmentation results from $p_{fg}(i,j)$. For example in Fig. \ref{fig:teaser}a, given an input expression ``the girl with red tie'', the method above should be able to output an image mask covering the two persons. Although this method cannot separate individual object instances of the same class, the results can serve as a coarse segmentation for the expression, and be integrated into to the LSTM-CNN \cite{hu2016segmentation} to reduce category errors. 

In our implementation, we use the Fully Convolutional Network (FCN) model \cite{long2015fully} as the category-based image segmentation model.

\subsection{Full Model}
\label{sec:all-in-1}

We combine the approaches described above for final output by taking a weighted average of the per-pixel foreground probability outputs $p_{fg}^{(1)}$ and $p_{fg}^{(2)}$ from the LSTM-CNN model (augmented with word embeddings and synthesized expressions in Sec. \ref{sec:method_1} and \ref{sec:syn_expr}) and category-based image segmentation model in Sec. \ref{sec:method_2}, as follows
\begin{equation}
p_{fg}(i,j) = \alpha \cdot p_{fg}^{(1)}(i,j) + (1-\alpha) \cdot p_{fg}^{(2)}(i,j)
\label{eqn:all-in-1}
\end{equation}
where $\alpha \in (0, 1)$ determines the weight of the two outputs. $\alpha$ is trained jointly with the whole system, end-to-end with back-propagation.

\section{Experiments}

\begin{table*}[t]
\scriptsize
\begin{center}
\begin{tabular}{|l|r|r|r|r|r||r|}
\hline
\textbf{Method} & \textit{prec}@0.5 & \textit{prec}@0.6 & \textit{prec}@0.7 & \textit{prec}@0.8 & \textit{prec}@0.9 & overall IoU \\
\hline
baseline LSTM-CNN \cite{hu2016segmentation}
& 15.25\% &  8.37\% &  3.75\% &  1.29\% &  0.06\% & 28.14\% \\
ours (with word embedding)
& 16.44\% &  9.25\% &  4.35\% &  1.39\% &  0.04\% & 30.72\% \\
ours (with word embedding, synthesized expressions)
& 17.38\% & 10.40\% &  4.72\% &  1.48\% &  0.07\% & 31.52\% \\
ours (with category-base image segmentation)
& 20.89\% & 13.07\% &  6.65\% &  2.73\% &  0.36\% & 33.53\% \\
\hline
ours (full model)
& \textbf{21.08\%} & \textbf{13.34\%} &  \textbf{7.47\%} &  \textbf{2.98\%} &  \textbf{0.44\%} & \textbf{34.06\%} \\
\hline
\end{tabular}
\end{center}
\caption{G-Ref dataset: The precision and overall IoU of our methods and baseline approach (higher is better). }
\label{tab:results_refgoog}

\scriptsize
\begin{center}
\begin{tabular}{|l|r|r|r|r|r||r|}
\hline
\textbf{Method} & \textit{prec}@0.5 & \textit{prec}@0.6 & \textit{prec}@0.7 & \textit{prec}@0.8 & \textit{prec}@0.9 & overall IoU \\
\hline
baseline LSTM-CNN \cite{hu2016segmentation}
& 26.82\% & 16.04\% &  7.58\% &  1.83\% &  0.06\% & 35.34\% \\
ours (with word embedding)
& 26.80\% & 16.26\% &  7.52\% &  1.91\% &  0.05\% & 35.24\% \\
ours (with word embedding, synthesized expressions)
& 26.92\% & 16.57\% &  8.14\% &  2.04\% &  0.01\% & 35.53\% \\
ours (with category-base image segmentation)
& 10.38\% &  5.27\% &  2.21\% &  0.72\% &  0.05\% & 26.10\% \\
\hline
ours (full model)
& \textbf{27.56\%} & \textbf{17.06\%} &  \textbf{8.18\%} &  \textbf{2.23\%} &  \textbf{0.13\%} & \textbf{36.05\%} \\
\hline
\end{tabular}
\end{center}
\caption{UNC-Ref dataset: The precision and overall IoU of our methods and baseline approach (higher is better). }
\label{tab:results_refcoco}

\scriptsize
\begin{center}
\begin{tabular}{|l|r|r|r|r|r||r|}
\hline
\textbf{Method} & \textit{prec}@0.5 & \textit{prec}@0.6 & \textit{prec}@0.7 & \textit{prec}@0.8 & \textit{prec}@0.9 & overall IoU \\
\hline
baseline LSTM-CNN \cite{hu2016segmentation}
& 34.02\% & 26.71\% & 19.32\% & 11.63\% &  3.92\% & 48.03\% \\
ours (with word embedding)
& \textbf{35.86\%} & \textbf{28.37\%} & \textbf{20.49\%} & \textbf{12.47\%} &  \textbf{4.48\%} & \textbf{49.91\%} \\
\hline
\end{tabular}
\end{center}
\caption{ReferIt dataset: The precision and overall IoU of our methods and baseline approach (higher is better). }
\label{tab:results_refclef}
\end{table*}

We evaluate how additional vision-only and text-only datasets with our method can help improve the performance of image segmentation from referring expressions. Existing image segmentation datasets with joint vision and language annotations include G-Ref, UNC-Ref \cite{mao2016generation} and ReferIt \cite{kazemzadeh2014referitgame}, containing 25799, 19994 and 19997 images, respectively. Also, MS COCO \cite{lin2014microsoft} and Gigaword are used as additional vision-only and text-only datasets. MS COCO contains 123287 images with class label annotations over image regions (Fig. \ref{fig:teaser}b). Both G-Ref and UNC-Ref are built upon MS COCO and have image regions annotated with both visual class labels and textual expressions, while ReferIt contains image regions with textual expressions but no class labels.

\textbf{Baseline.} On G-Ref\footnote{We report the performance on G-Ref using its validation set, since its test set has not been released.}, UNC-Ref and ReferIt, we train and evaluate the LSTM-CNN baseline under precision metric and overall IoU metrics \cite{hu2016segmentation}, shown in Tables \ref{tab:results_refgoog}, \ref{tab:results_refcoco}, and \ref{tab:results_refclef}.

\textbf{Word embedding and synthesized expressions.} As described in Sec. \ref{sec:method_1}, we use the GloVe vectors pretrained on Gigaword as our word embedding matrix, and add image regions with synthesized expressions as additional training data (Sec. \ref{sec:syn_expr}). Tables \ref{tab:results_refgoog}, \ref{tab:results_refcoco}, and \ref{tab:results_refclef} show that the pretrained word embedding and additional synthesized data improves the image segmentation performance over the baseline.

\begin{figure}[t]
\centering
\scriptsize
\begin{tabularx}{0.95\linewidth}{YYYY}
input image & $p_{fg}$ heatmap & final output & ground-truth \\ \hline
\end{tabularx} \\
\scriptsize{(a) input expression=\refexp{bird on right side of windowsill}} \\
\includegraphics[trim = 56mm 54mm 45mm 51mm, clip=true,width=0.9\linewidth,height=\textheight,keepaspectratio]{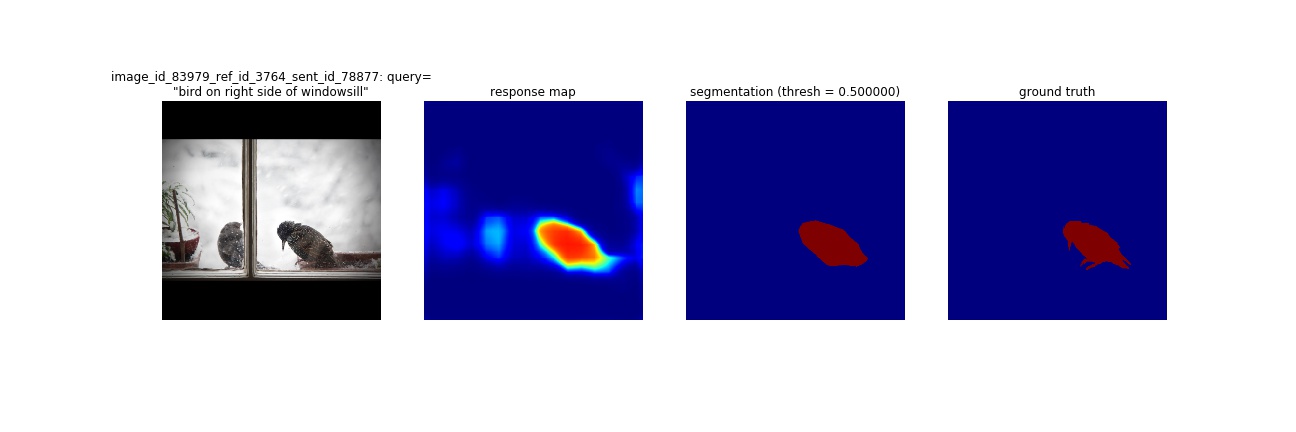} \\
\scriptsize{(b) input expression=\refexp{a cell phone cover which was opened to repair}} \\
\includegraphics[trim = 56mm 51mm 45mm 48mm, clip=true,width=0.9\linewidth,height=\textheight,keepaspectratio]{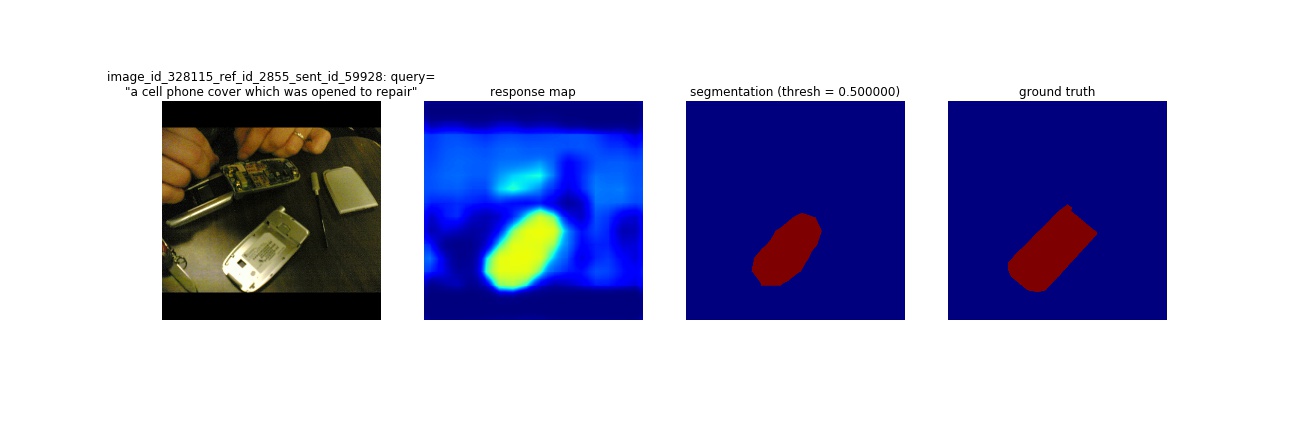} \\
\scriptsize{(c) input expression=\refexp{screen in middle facing you}} \\
\includegraphics[trim = 56mm 51mm 45mm 48mm, clip=true,width=0.9\linewidth,height=\textheight,keepaspectratio]{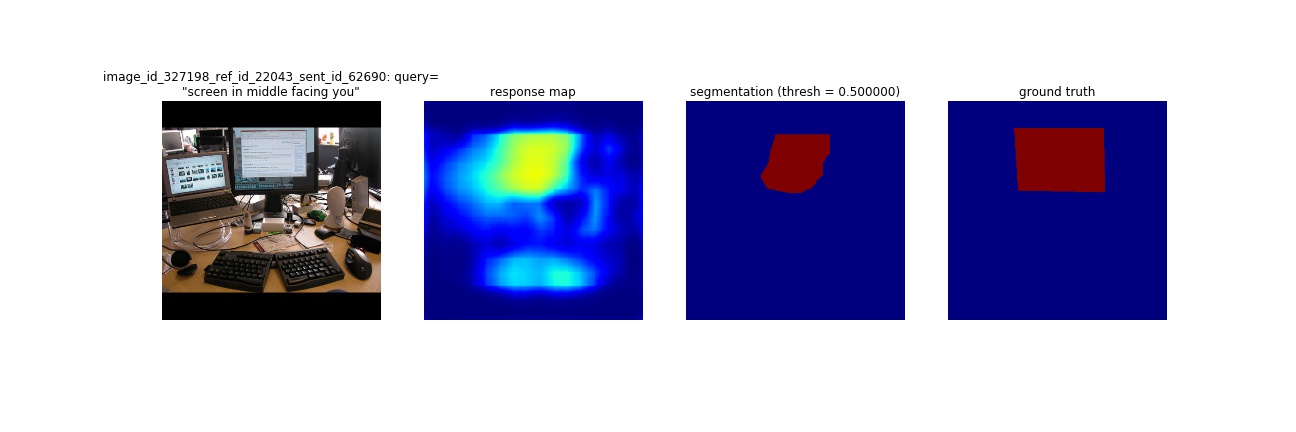} \\
\scriptsize{(d) input expression=\refexp{left kid}} \\
\includegraphics[trim = 56mm 54mm 45mm 51mm, clip=true,width=0.9\linewidth,height=\textheight,keepaspectratio]{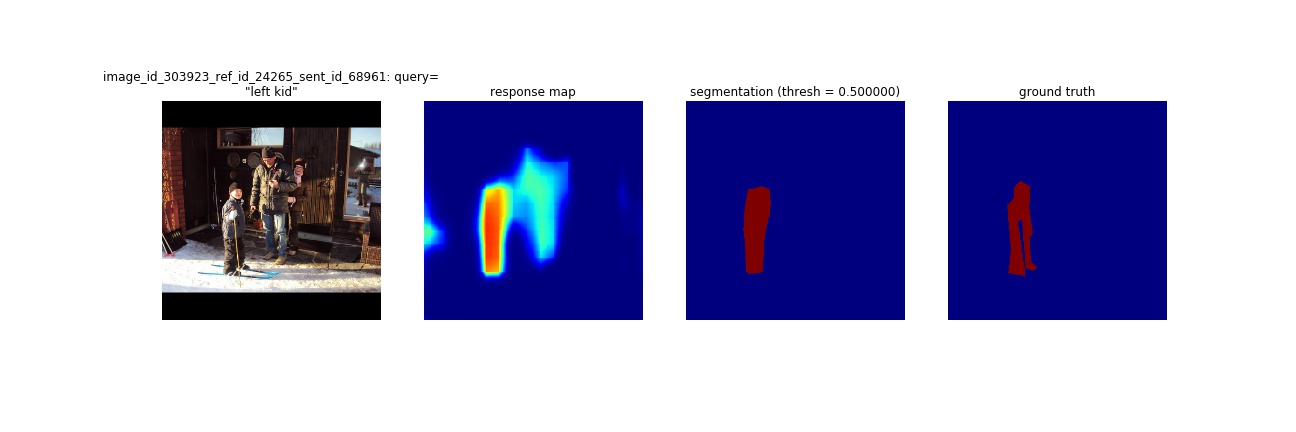} \\
\caption{Example results on G-Ref (a, b) and UNC-Ref (c, d) with our full model. }
\label{fig:visualized_results}
\end{figure}

\textbf{Category-based image segmentation and full model.} As described in Sec. \ref{sec:method_2}, we train visual category classifiers on G-Ref and UNC-Ref for the textual expressions, and train a FCN model on MS COCO for category-based image segmentation, and obtain the full model as in Sec. \ref{sec:all-in-1}. Fig. \ref{fig:visualized_results} shows some visualized results from our full model and Table \ref{tab:results_refgoog} and \ref{tab:results_refcoco} show the performance of category-based image segmentation and the full model on G-Ref and UNC-Ref. It can be seen that our full model achieves the highest performance, outperforming previous results.

\begin{figure*}
\centering
\small
\vspace{0.3 cm}
\begin{tabularx}{0.95\textwidth}{YYYY}
input image & ours (with word embedding, synthesized expressions & ours (with category-base image segmentation) & ours (full model) \\ \hline
\end{tabularx} \\
\vspace{0.3 cm}
\small{(b) input expression=\refexp{a laptop open with multiple windows opened on it}} \\
\includegraphics[trim = 56mm 49mm 45mm 46mm, clip=true,width=0.9\textwidth,height=\textheight,keepaspectratio]{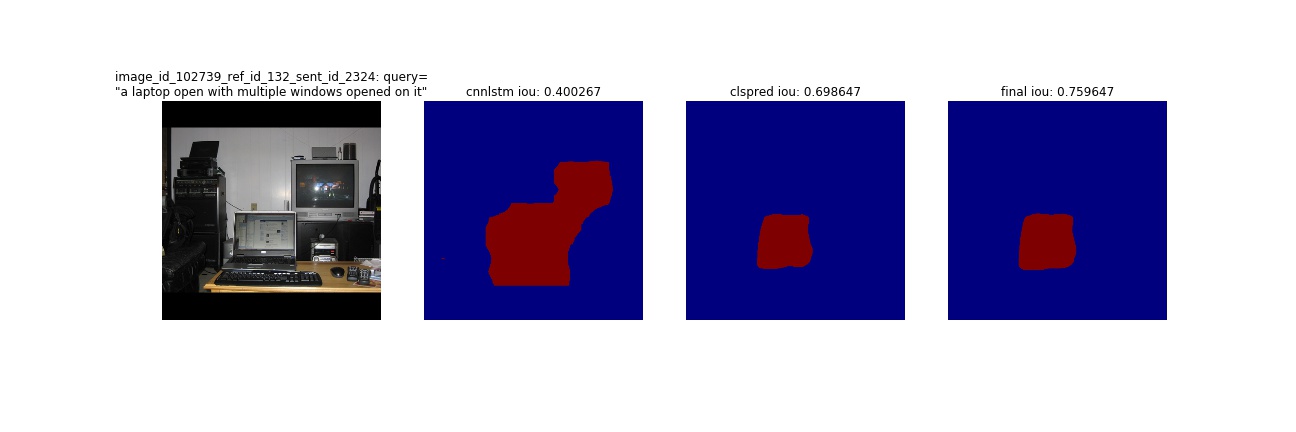} \\
\vspace{0.2 cm}
\small{(a) input expression=\refexp{camouflauged plane}} \\
\includegraphics[trim = 56mm 52mm 45mm 48mm, clip=true,width=0.9\textwidth,height=\textheight,keepaspectratio]{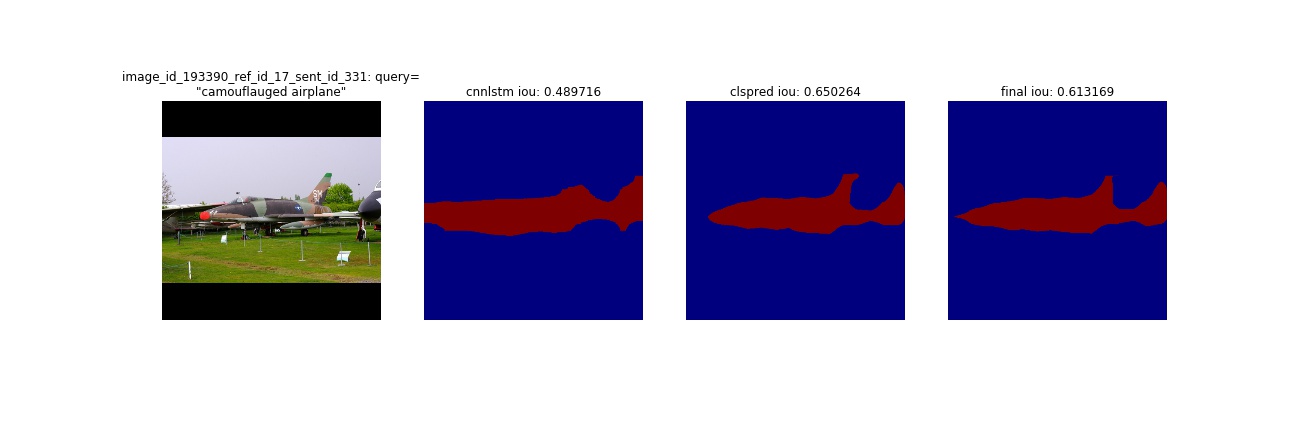} \\
\vspace{0.2 cm}
\small{(c) input expression=\refexp{red umbrella with woman in black dotted shirt sitting under it}} \\
\includegraphics[trim = 56mm 52mm 45mm 48mm, clip=true,width=0.9\textwidth,height=\textheight,keepaspectratio]{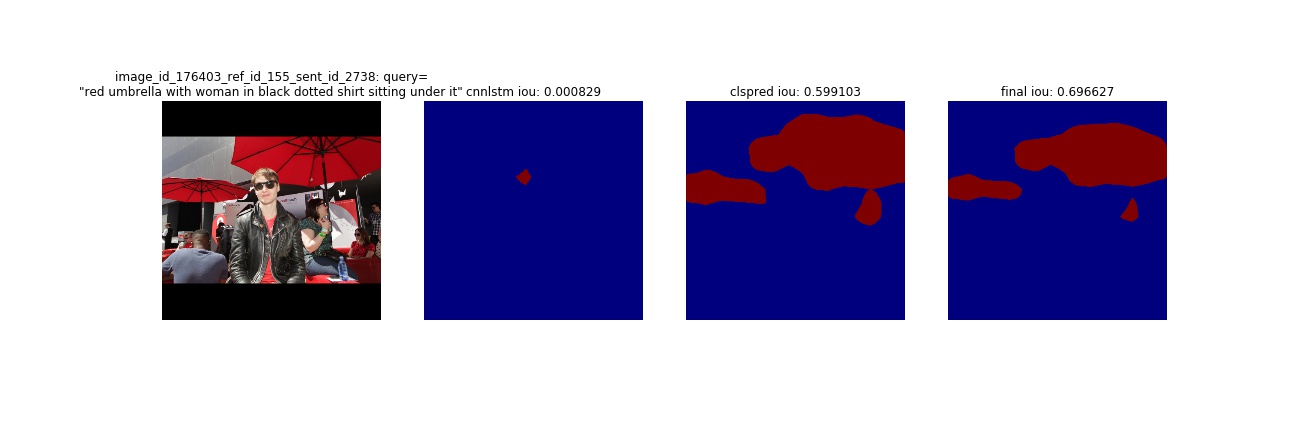} \\
\vspace{0.6 cm}
\small{(d) input expression=\refexp{banana bunch at top}} \\
\includegraphics[trim = 56mm 38mm 45mm 35mm, clip=true,width=0.9\textwidth,height=\textheight,keepaspectratio]{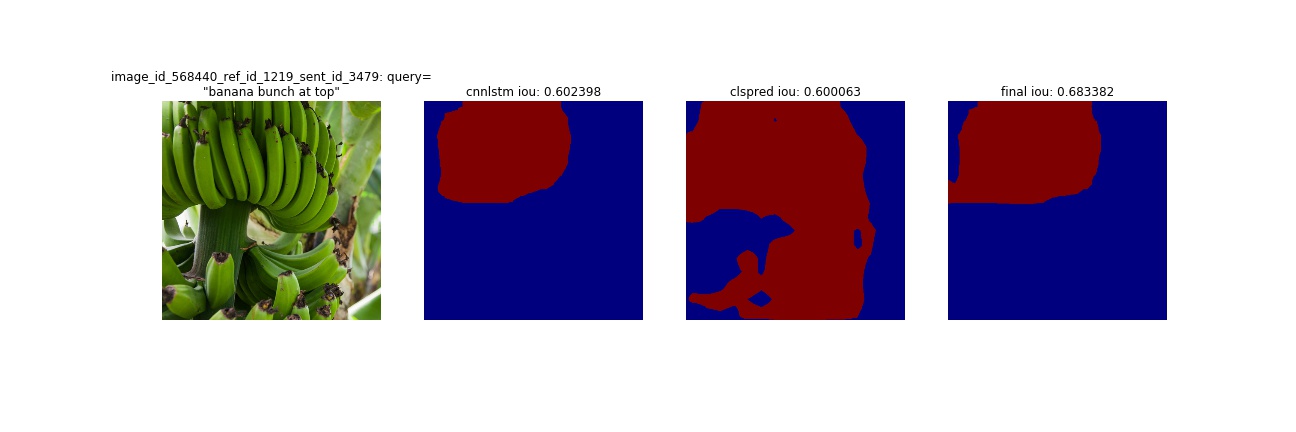} \\
\vspace{0.2 cm}
\small{(e) input expression=\refexp{elephant in front of you}} \\
\includegraphics[trim = 56mm 51mm 45mm 48mm, clip=true,width=0.9\textwidth,height=\textheight,keepaspectratio]{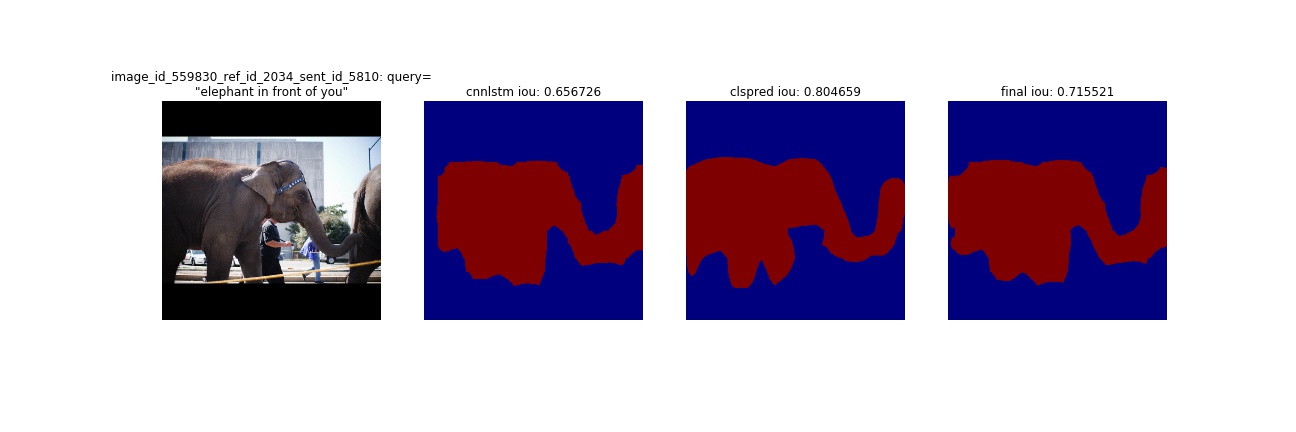} \\
\caption{Visualized comparisons on G-Ref (shown in a, b, c) and UNC-Ref (shown in d, e) of different segmentation outputs from different components in our method. The second to last columns correspond to the results in Table 1 and 2 in the main paper.}
\label{fig:results_supp_comp}
\end{figure*}

\begin{figure*}
\centering
\small
\begin{tabularx}{0.95\textwidth}{YYYY}
input image & $p_{fg}$ heatmap & final output & ground-truth \\ \hline
\end{tabularx} \\
\small{input expression=\refexp{a bowl of noodles with some broccoli}} \\
\includegraphics[trim = 56mm 38mm 45mm 35mm, clip=true,width=0.9\textwidth,height=\textheight,keepaspectratio]{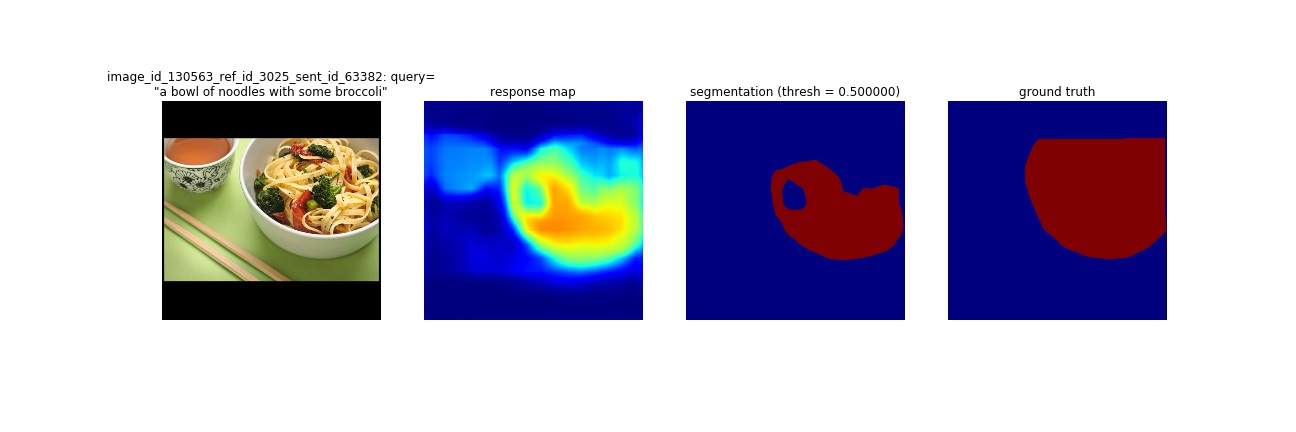} \\
\small{input expression=\refexp{this is the larger screen of a computer . also a smaller one there}} \\
\includegraphics[trim = 56mm 38mm 45mm 35mm, clip=true,width=0.9\textwidth,height=\textheight,keepaspectratio]{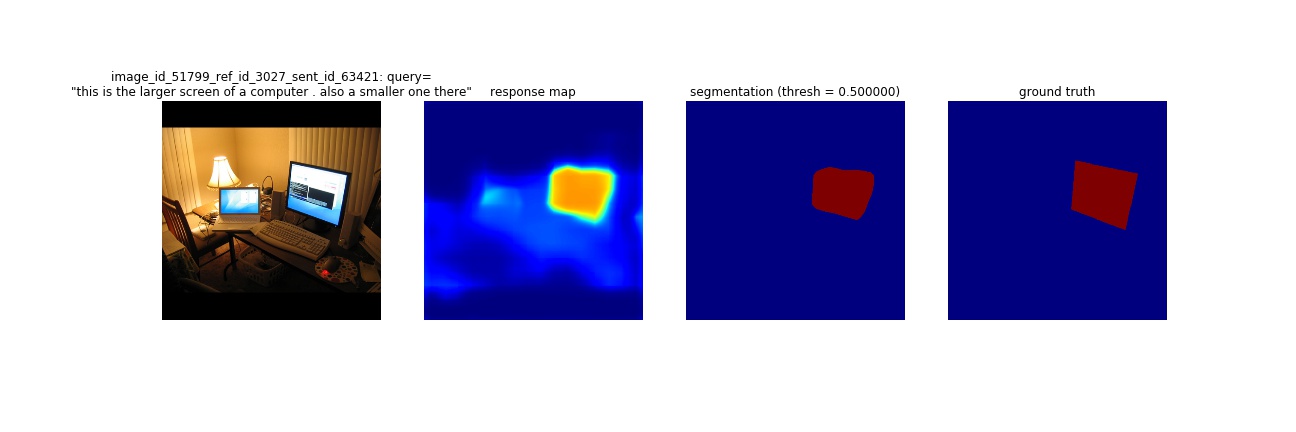} \\
\small{input expression=\refexp{a small bowl of fruit}} \\
\includegraphics[trim = 56mm 38mm 45mm 35mm, clip=true,width=0.9\textwidth,height=\textheight,keepaspectratio]{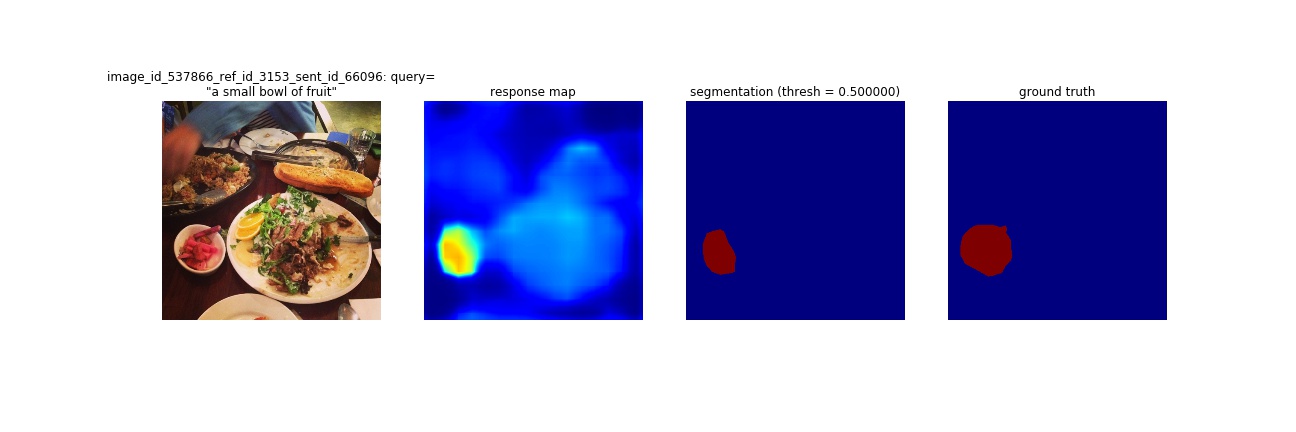} \\
\small{input expression=\refexp{the traffic light}} \\
\includegraphics[trim = 56mm 38mm 45mm 35mm, clip=true,width=0.9\textwidth,height=\textheight,keepaspectratio]{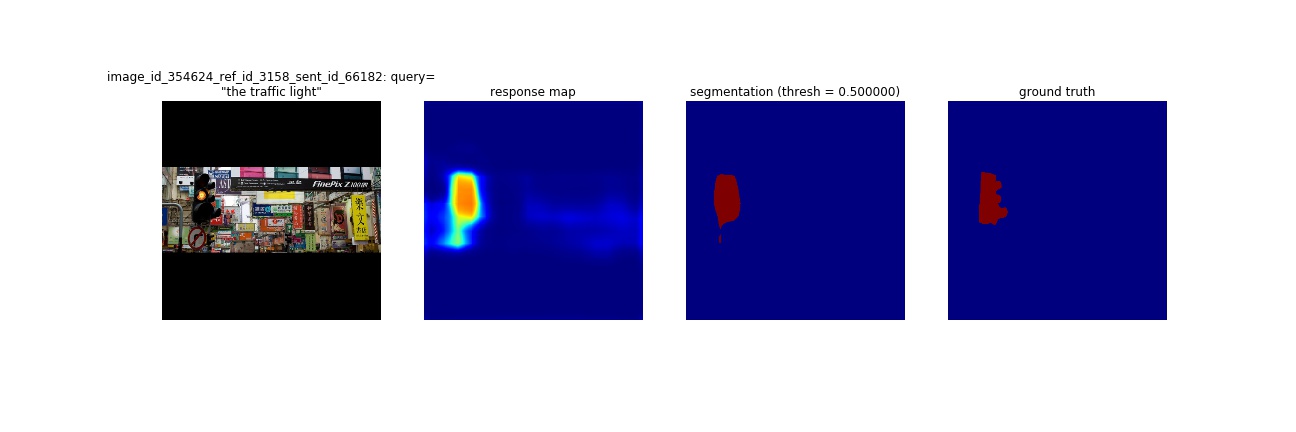} \\
\small{input expression=\refexp{a black and white clock at 12 : 00 on the side of a large tower}} \\
\includegraphics[trim = 56mm 38mm 45mm 35mm, clip=true,width=0.9\textwidth,height=\textheight,keepaspectratio]{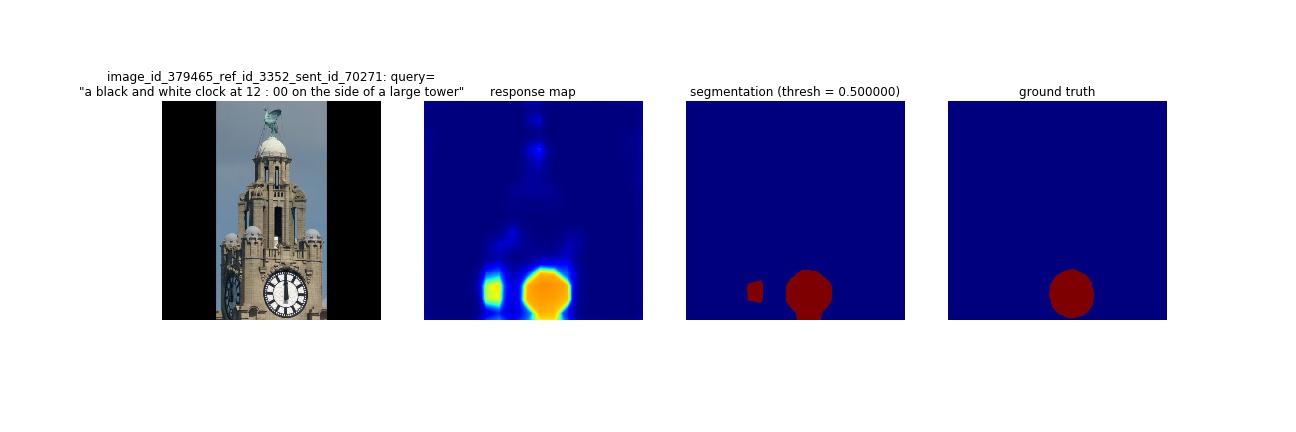} \\
\small{input expression=\refexp{the umbrella is pink and orange}} \\
\includegraphics[trim = 56mm 38mm 45mm 35mm, clip=true,width=0.9\textwidth,height=\textheight,keepaspectratio]{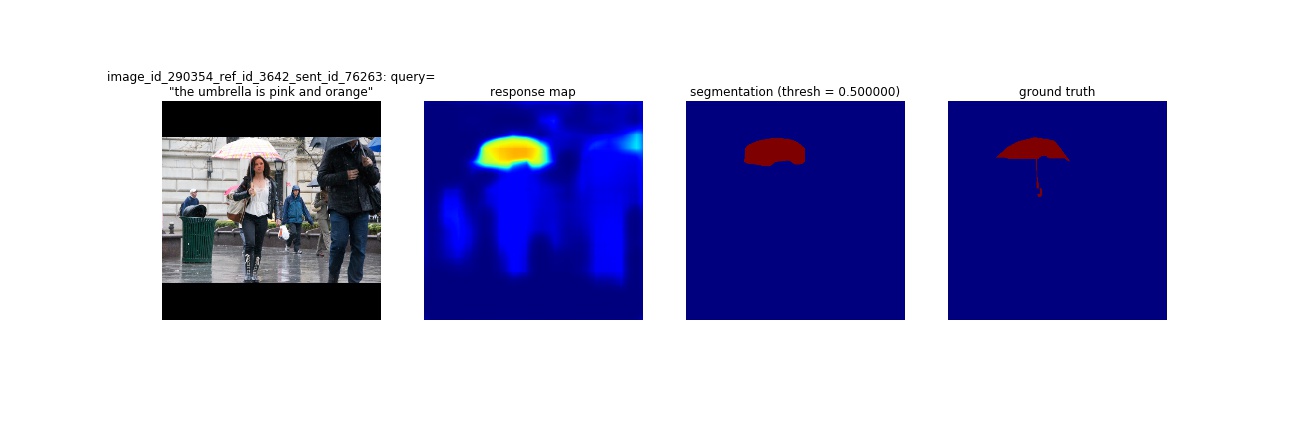} \\
\caption{Example image segmentation results on G-Ref with our full model. }
\label{fig:results_supp_refgoog}
\end{figure*}

\begin{figure*}
\centering
\small
\begin{tabularx}{0.95\textwidth}{YYYY}
input image & $p_{fg}$ heatmap & final output & ground-truth \\ \hline
\end{tabularx} \\
\small{input expression=\refexp{girl in hat}} \\
\includegraphics[trim = 56mm 38mm 45mm 35mm, clip=true,width=0.9\textwidth,height=\textheight,keepaspectratio]{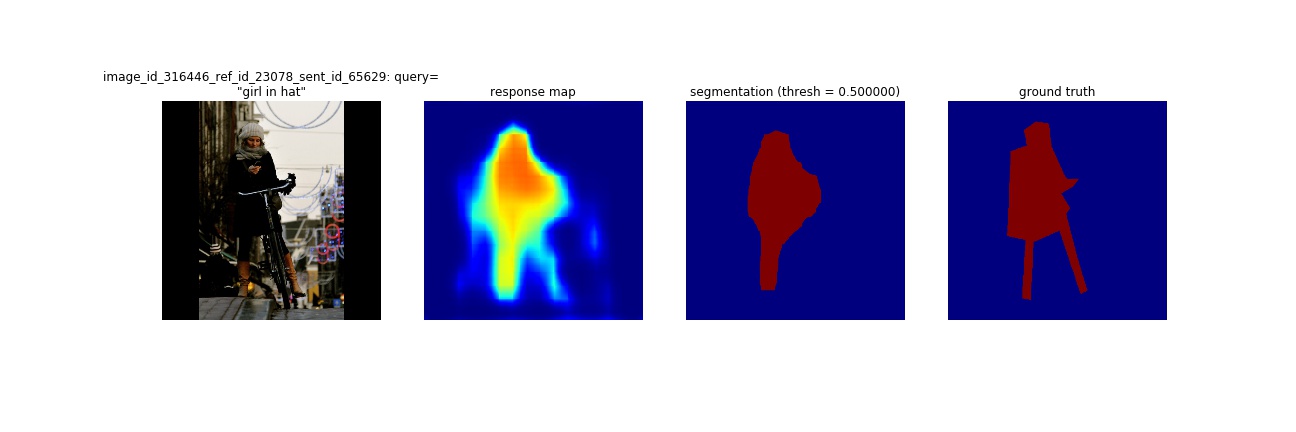} \\
\small{input expression=\refexp{guy on skate board in black coat}} \\
\includegraphics[trim = 56mm 38mm 45mm 35mm, clip=true,width=0.9\textwidth,height=\textheight,keepaspectratio]{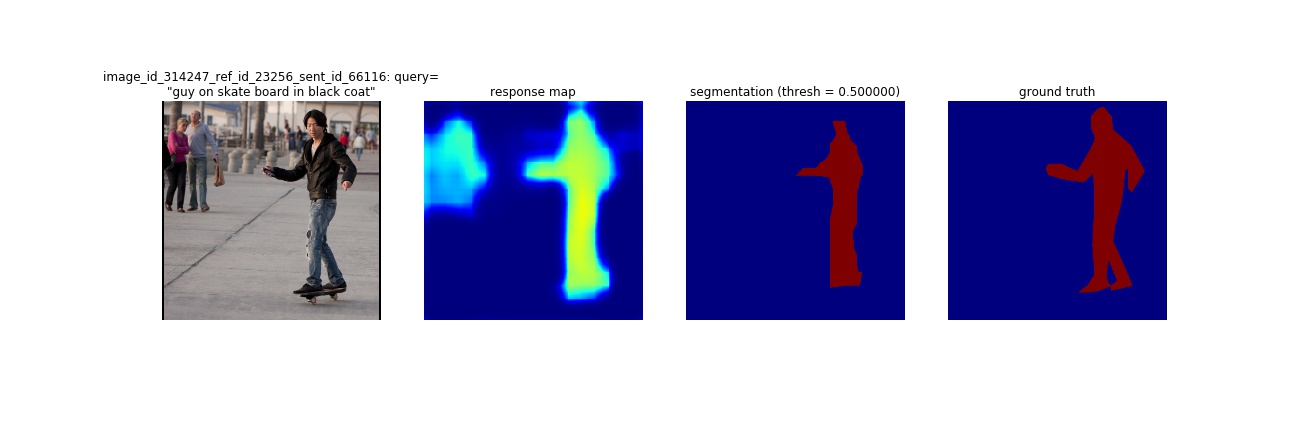} \\
\small{input expression=\refexp{the one on the left}} \\
\includegraphics[trim = 56mm 38mm 45mm 35mm, clip=true,width=0.9\textwidth,height=\textheight,keepaspectratio]{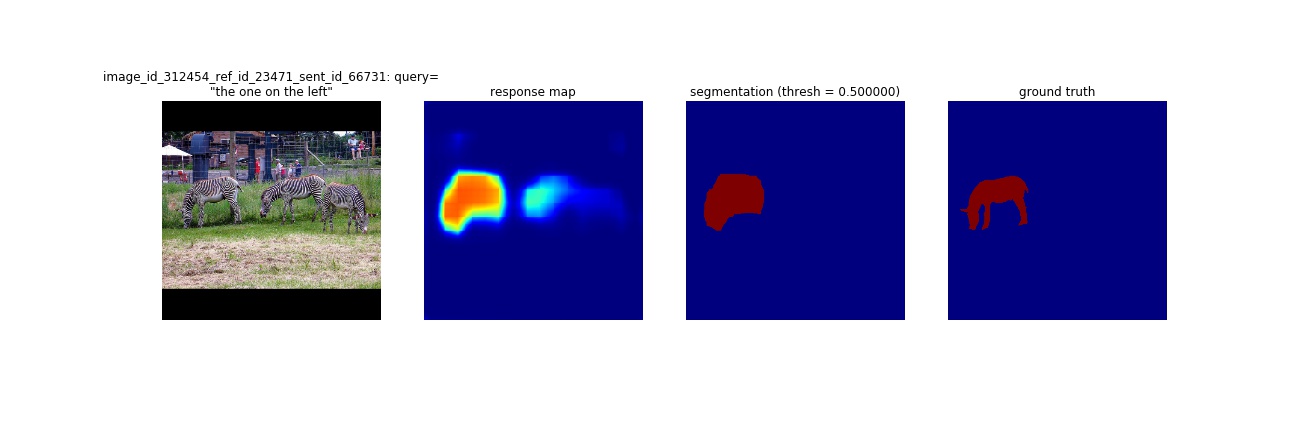} \\
\small{input expression=\refexp{hot dog left}} \\
\includegraphics[trim = 56mm 38mm 45mm 35mm, clip=true,width=0.9\textwidth,height=\textheight,keepaspectratio]{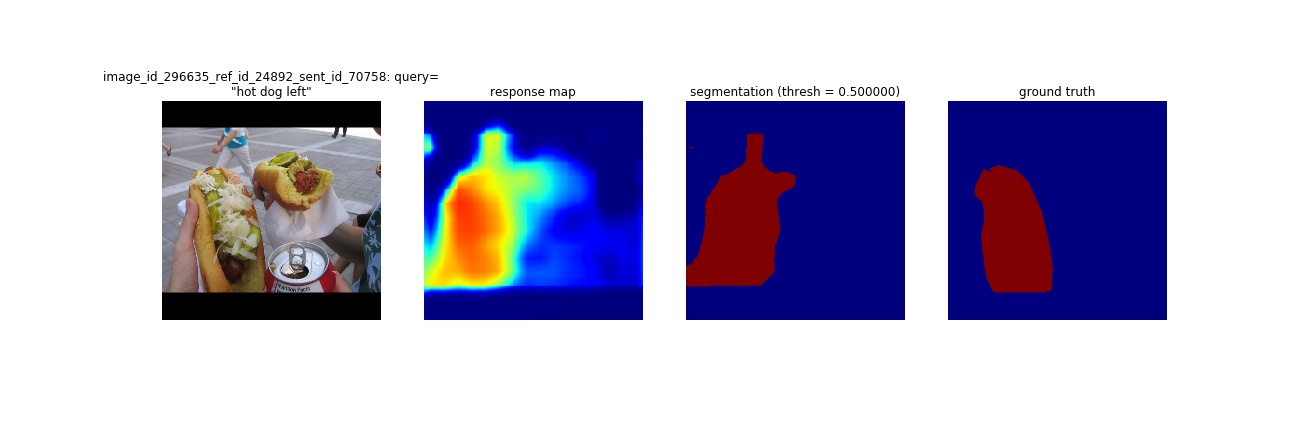} \\
\small{input expression=\refexp{female in front cooking on the right}} \\
\includegraphics[trim = 56mm 38mm 45mm 35mm, clip=true,width=0.9\textwidth,height=\textheight,keepaspectratio]{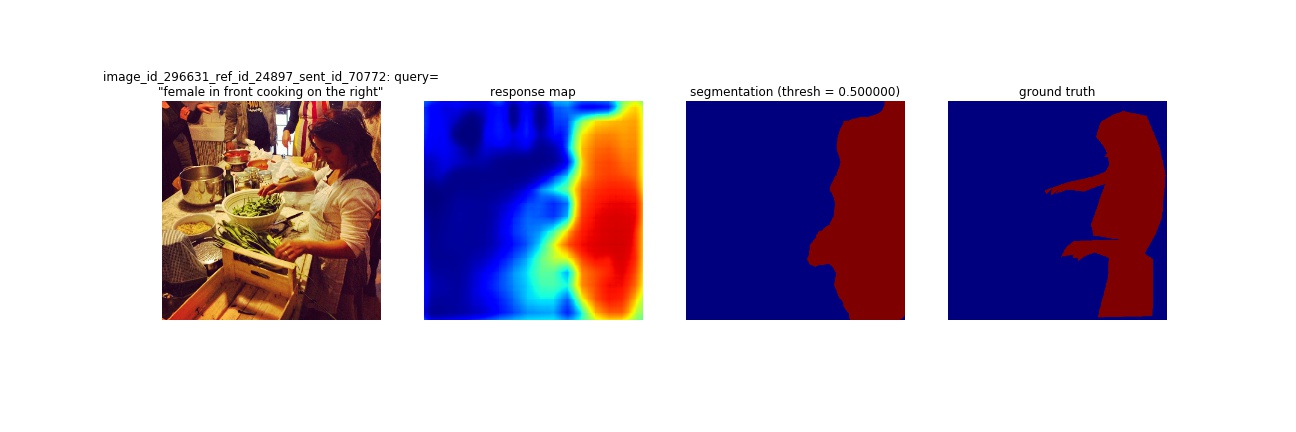} \\
\small{input expression=\refexp{left in green}} \\
\includegraphics[trim = 56mm 38mm 45mm 35mm, clip=true,width=0.9\textwidth,height=\textheight,keepaspectratio]{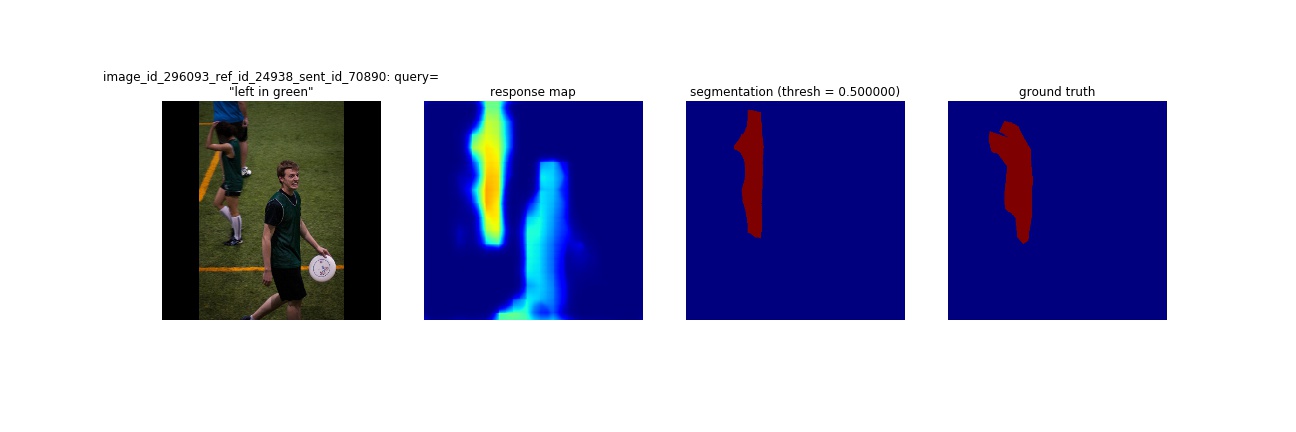} \\
\caption{Example image segmentation results on UNC-Ref with our full model. }
\label{fig:results_supp_refcoco}
\end{figure*}

\begin{figure*}
\centering
\small
\begin{tabularx}{0.95\textwidth}{YYYY}
input image & $p_{fg}$ heatmap & final output & ground-truth \\ \hline
\end{tabularx} \\
\small{input expression=\refexp{green stuff}} \\
\includegraphics[trim = 56mm 38mm 45mm 35mm, clip=true,width=0.9\textwidth,height=\textheight,keepaspectratio]{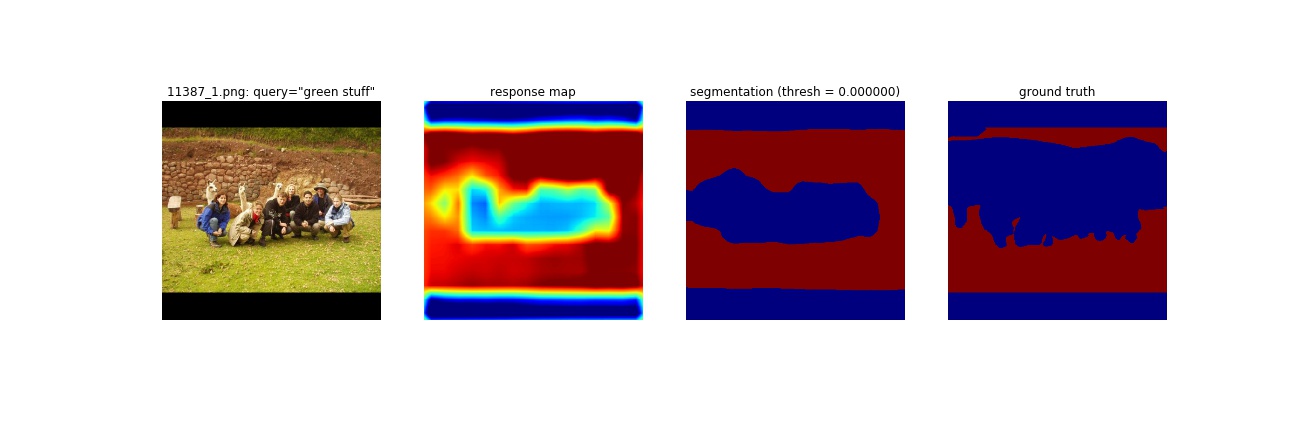} \\
\small{input expression=\refexp{tree leaves}} \\
\includegraphics[trim = 56mm 38mm 45mm 35mm, clip=true,width=0.9\textwidth,height=\textheight,keepaspectratio]{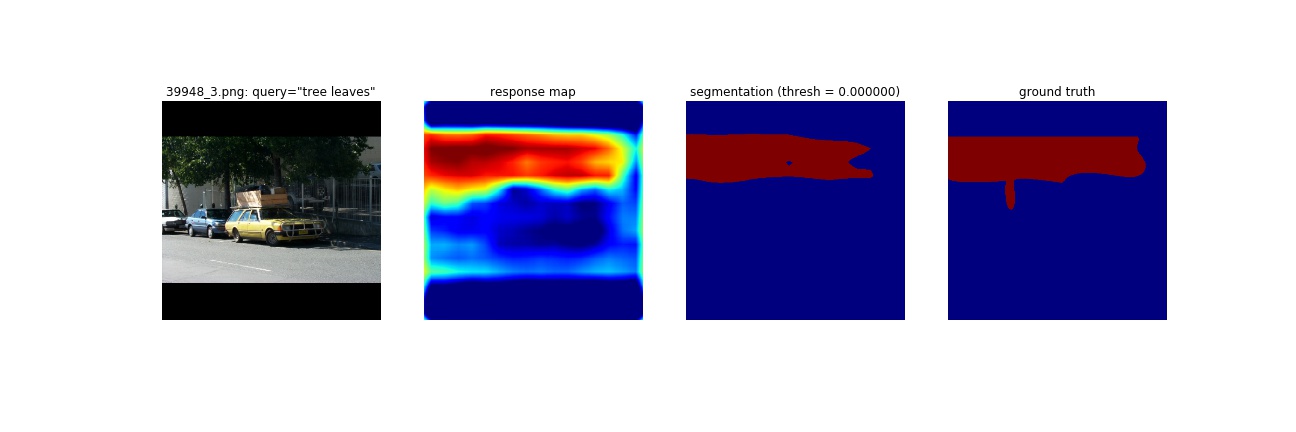} \\
\small{input expression=\refexp{roo}} \\
\includegraphics[trim = 56mm 38mm 45mm 35mm, clip=true,width=0.9\textwidth,height=\textheight,keepaspectratio]{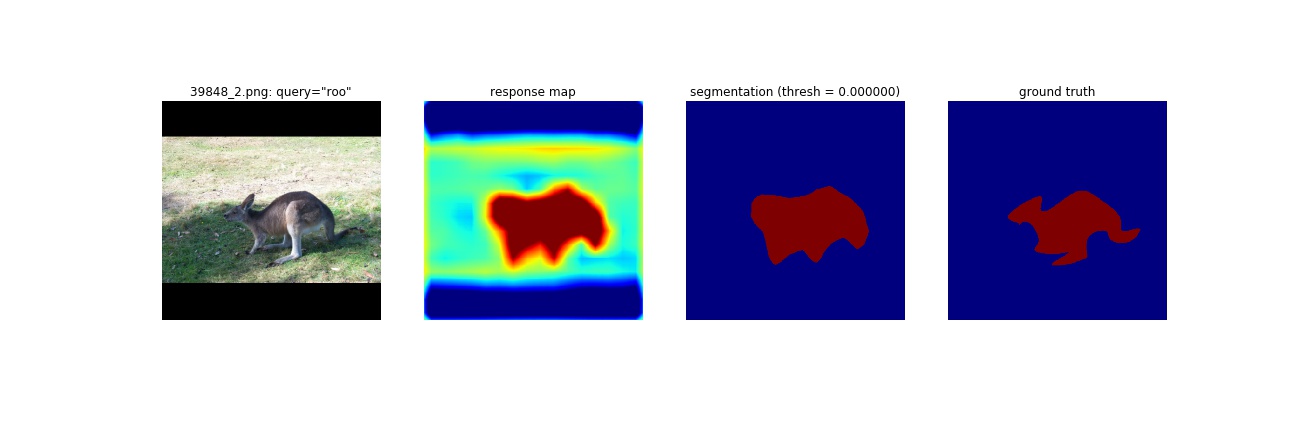} \\
\small{input expression=\refexp{tree to left}} \\
\includegraphics[trim = 56mm 38mm 45mm 35mm, clip=true,width=0.9\textwidth,height=\textheight,keepaspectratio]{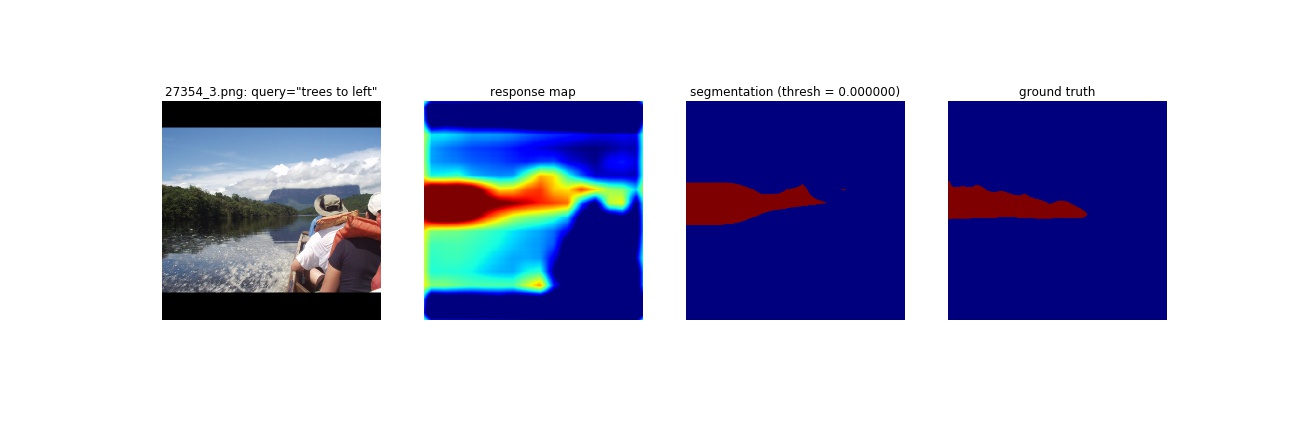} \\
\small{input expression=\refexp{water}} \\
\includegraphics[trim = 56mm 38mm 45mm 35mm, clip=true,width=0.9\textwidth,height=\textheight,keepaspectratio]{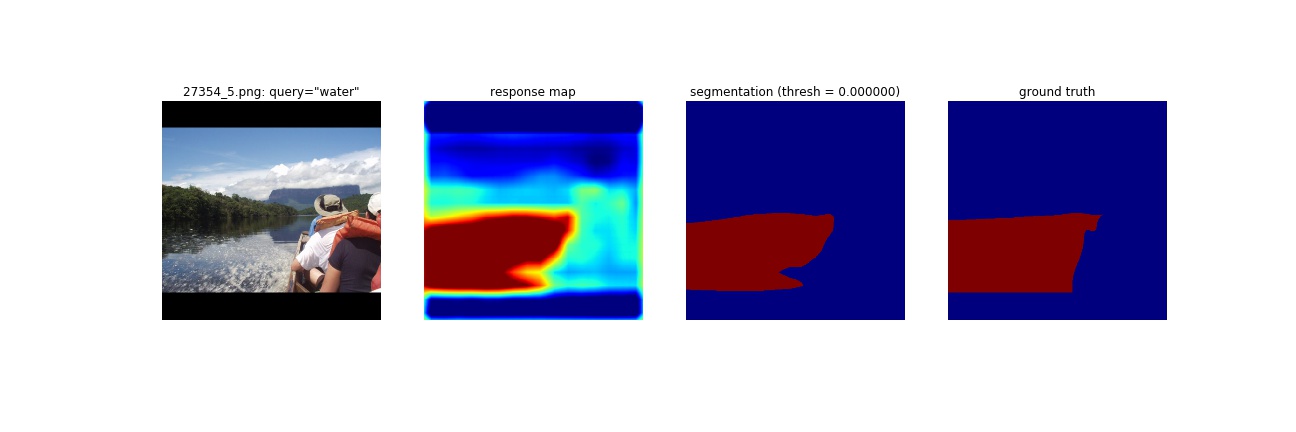} \\
\small{input expression=\refexp{people}} \\
\includegraphics[trim = 56mm 38mm 45mm 35mm, clip=true,width=0.9\textwidth,height=\textheight,keepaspectratio]{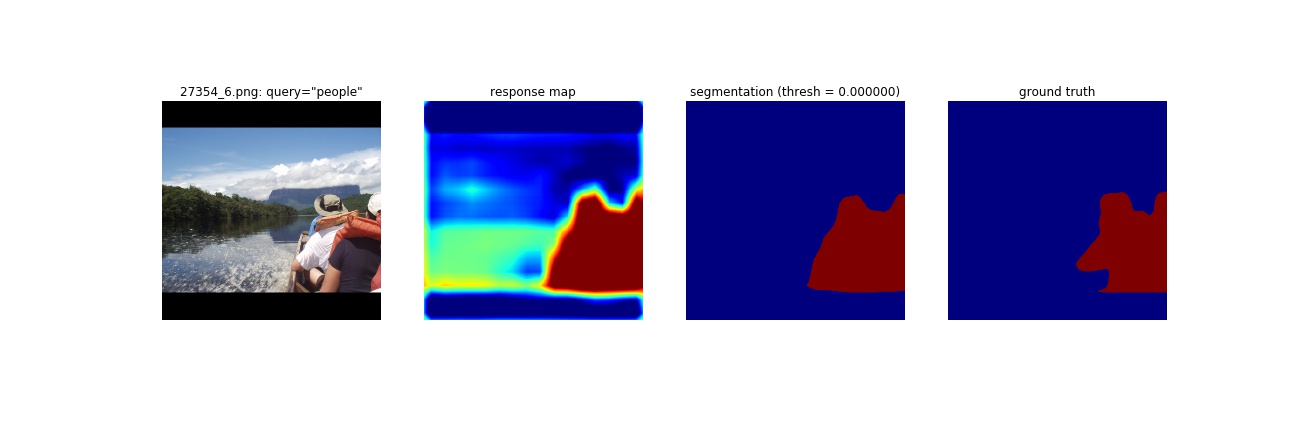} \\
\caption{Example image segmentation results on ReferIt with our model (with word embedding). }
\label{fig:results_supp_refclef}
\end{figure*}

Figure \ref{fig:results_supp_comp} visualizes image segmentation predictions using different components of our method, where the second to last columns correspond to the results in Table 1 and 2 in the paper. Figure \ref{fig:results_supp_refgoog}, \ref{fig:results_supp_refcoco} and \ref{fig:results_supp_refclef} contain more results on G-Ref, UNC-Ref and ReferIt dataset.

\section{Conclusion}

In this paper, we propose a method to utilize additional large scale vision and text data to improve the performance of image segmentation from referring expressions. A word embedding is pretrained from large corpora and textual expressions are synthesized from visual class labels as additional data. Also, well-studied traditional category-based semantic image segmentation models are integrated into language-based image segmentation. Experimental results show that the method in this paper improves the performance over previous work.

As future work, we would like to extend our method to incorporate recent datasets containing entities and bounding box annotations \cite{krishna2016visual,plummer2015flickr30k}.

\bibliography{biblioLong,references}
\bibliographystyle{emnlp2016}

\end{document}